\documentclass[runningheads]{llncs}
\usepackage{graphicx}
\usepackage{soul}
\usepackage{booktabs} % provide \toprule, \midrule, bottomrule
\usepackage[table,xcdraw]{xcolor}

\usepackage{amsfonts}
\usepackage[colorlinks,bookmarksopen,bookmarksnumbered,citecolor=blue, linkcolor=blue, urlcolor=blue]{hyperref}

\usepackage{amsmath}
\usepackage{subfigure}

\usepackage{wrapfig} % 提供 wrapfigure 环境

\usepackage{lipsum} % 用于生成随机文本
\usepackage{caption} % 添加 caption 包
\usepackage{diagbox}
\usepackage{array}
\usepackage{manyfoot}
\usepackage[misc]{ifsym} 

\hypersetup{
colorlinks=true,
linkcolor=black,
citecolor=blue,
urlcolor=magenta  
}

\begin{document}
%
% \title{Visual Object Tracking under Transformations: A Benchmark}
\title{Tracking Transforming Objects: A Benchmark}
% \author{Paper ID: 1452 }
\author{
You Wu\inst{1,\dagger}\and
Yuelong Wang\inst{1,{\dagger}}\and
Yaxin Liao\inst{1}
\and
Fuliang Wu\inst{1} 
\and
Hengzhou Ye\inst{1,2}
\and
Shuiwang Li\inst{1,2,{(\textrm{\Letter})}}
}
% \authorrunning{Paper ID: 1452 et al.}
\authorrunning{Y. Wu et al.}
\institute{Guilin University of Technology, Guilin, China \and
Guangxi Key Laboratory of Embedded Technology and Intelligent System, Guilin, China\\
\email{lishuiwang0721@163.com} \\
% \url{https://github.com/wuyou3474/DTTO}
}
\maketitle              % typeset the header of the contribution
\begin{abstract}
% During an object's transformation, its appearance might be transient.
% For instance, transformers may transition between robot and car forms, while flowers can transition from a bud to a fully blooming state. 
% In these process, their color, shape, and texture can undergo significant transformations, often maintaining little resemblance to their original form yet still retaining their inherent identity.
% However, this significant phenomenon has not garnered adequate attention in current visual object tracking benchmark.
% In this study, we bridge this gap by collecting a novel dedicated \textbf{D}ataset for \textbf{T}racking \textbf{T}ransforming \textbf{O}bjects, called DTTO.
% DTTO contains 100 sequences, amounting to approximately 9.3K frames.
% We provide carefully hand-annotated bounding boxes for each frame within these sequences, making DTTO the pioneering benchmark dedicated to tracking transforming objects.
% We thoroughly evaluate 20 state-of-the-art trackers on the benchmark, aiming to comprehend the performance of existing methods and provide a comparison for future research on DTTO.
% With the release of DTTO, our goal is to facilitate further research and applications related to tracking transforming objects.

Tracking transforming objects holds significant importance in various fields due to the dynamic nature of many real-world scenarios. By enabling systems accurately represent transforming objects over time, tracking transforming objects facilitates advancements in areas such as autonomous systems, human-computer interaction, and security applications. Moreover, understanding the behavior of transforming objects provides valuable insights into complex interactions or processes, contributing to the development of intelligent systems capable of robust and adaptive perception in dynamic environments. However, current research in the field mainly focuses on tracking generic objects. In this study, we bridge this gap by collecting a novel dedicated \textbf{D}ataset for \textbf{T}racking \textbf{T}ransforming \textbf{O}bjects, called DTTO, which contains 100 sequences, amounting to approximately 9.3K frames. We provide carefully hand-annotated bounding boxes for each frame within these sequences, making DTTO the pioneering benchmark dedicated to tracking transforming objects.
We thoroughly evaluate 20 state-of-the-art trackers on the benchmark, aiming to comprehend the performance of existing methods and provide a comparison for future research on DTTO. With the release of DTTO, our goal is to facilitate further research and applications related to tracking transforming objects. The dataset and evaluation toolkit are available at \url{https://github.com/wuyou3474/DTTO}.

\end{abstract}
\section{Introduction}

\footnotetext[0]{$\dagger$ These authors contributed equally to this work. \\
Thanks to the support by Guangxi Science and Technology Base and Talent Special Project (No. Guike AD22035127).}

Object tracking stands as one of the cornerstone challenges in computer vision, finding application across diverse fields such as video surveillance, robotics, human-machine interaction, and motion analysis \cite{cui2022mixformer,Li2020AsymmetricDC,Li2021LearningRC,LI2022108614,ye2022joint,gao2023generalized,li2023adaptive,zeng2023towards,wang2023learning}. The advent of deep learning has notably propelled significant advancements in object tracking in recent years \cite{zhang2020ocean,guo2020siamcar,danelljan2020probabilistic,yan2021learning,Jiao2021DeepLI,wu2023dropmae,shi2024explicit,xie2024autoregressive,lilearning,wang2024enhancing}. However, despite substantial progress in the field, current research predominantly concentrates on tracking generic objects, with limited attention directed towards tracking transforming objects. Although exploratory studies have been conducted on segmenting transforming objects \cite{tokmakov2023breaking,yu2023video,goyal2023m3t} in video object segmentation tasks very recently, their objective is primarily to predict pixel-wise masks of objects undergoing transformations in each frame of the video, utilizing the initial frame mask as a reference. Moreover, these studies maintain the object category consistent throughout each sequence.
In this work, we investigate the tracking of transforming objects, which not only undergo apparent changes in appearance, context, or viewpoint but also may change categories during the transformation process. This presents a more conceptually challenging task due to the involvement of more complex cognitive processes \cite{malim1994cognitive,estes2022handbook,busemeyer2010cognitive}.
% \cite{malim1994cognitive,estes2022handbook,von1995cognitive,busemeyer2010cognitive}.
For instance, humans possess the ability to maintain the consistent identity of objects despite alterations in appearance, context, or viewpoint. This cognitive ability, known as object constancy, enables individuals to perceive an object's underlying identity regardless of variations in its form or visual representation \cite{thatcher2021foundations,estes2022handbook}. Moreover, humans demonstrate cognitive flexibility, allowing them to adjust their perceptions and comprehension based on new information or experiences \cite{malim1994cognitive,kellogg2003cognitive,busemeyer2010cognitive}. This adaptability enables individuals to recognize an object's identity even as it undergoes changes in form, such as witnessing a robot transform into a car within the context of a movie. These cognitive processes are essential for humans to successfully track transforming objects, yet none of the mainstream discriminative and generative methods take these into account. This study introduces the task of tracking transforming objects by predicting the bounding box for the transforming object in each frame of the video, which explicitly involves these cognitive processes and presents more conceptual challenges to tracking algorithms.

\begin{figure}[ht]
    \centering
    \begin{minipage}[b]{1.0\textwidth}
        \centering
        \includegraphics[width=\textwidth]{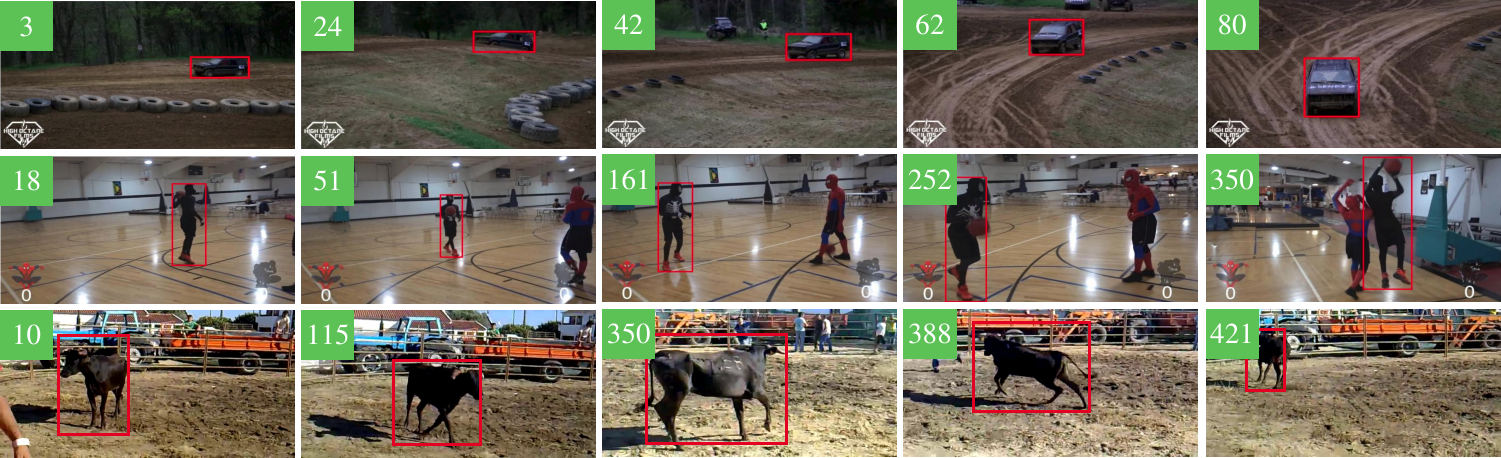}
        (a) Example of generic object tracking.
        \label{fig:sub1}
    \end{minipage}
    \hfill
    \begin{minipage}[b]{1.0\textwidth}
        \centering
        \includegraphics[width=\textwidth]{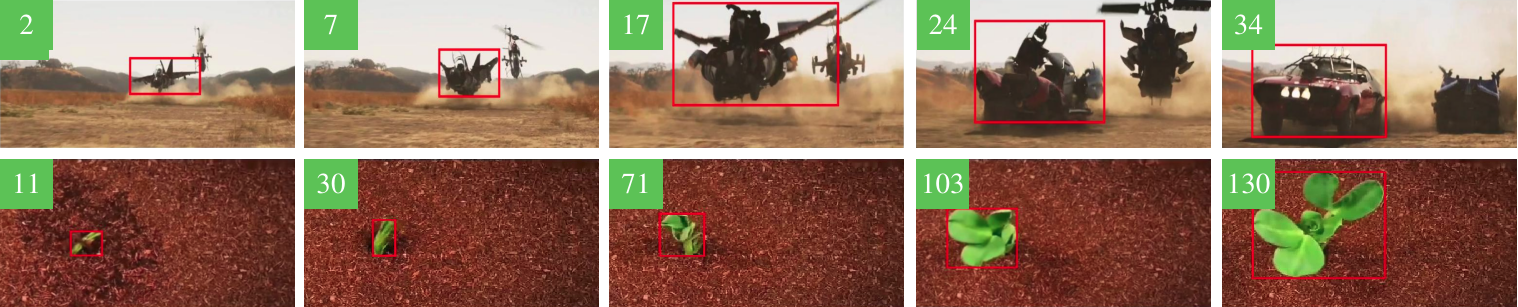}
        (b) Example of tracking transforming object.
        \label{fig:sub2}
    \end{minipage}
    \caption{The examples of generic object tracking from GOT-10k \cite{huang2019got}, LaSOT \cite{fan2019lasot}, and TrackingNet \cite{muller2018trackingnet} (a) and tracking transforming objects from our DTTO (b).}
    \label{fig:main}
\end{figure}

To support research on tracking transforming objects, in this paper we present a dedicated benchmark, called DTTO, which consists of 100 sequences, totaling approximately 9.3K frames, presenting six common transformation processes. Each sequence showcases significant transformations undergone by the object.
We provide carefully hand-annotated bounding boxes for each frame within these sequences.
To the best of our knowledge, DTTO is the first benchmark dedicated to tracking transforming objects, which aims to reveal the limitations of current visual object tracking (VOT) methods and identify the primary challenges in tracking transforming objects.
We conduct evaluations on this benchmark using 20 state-of-the-art (SOTA) trackers, aiming to comprehend the performance of existing methods and provide a comparison for future research on DTTO.
Fig. \ref{fig:main} presents the distinctions between generic object tracking and transforming object tracking proposed in this work, showing examples from Got-10k \cite{huang2019got}, LaSOT \cite{fan2019lasot}, TrackingNet    
 \cite{muller2018trackingnet} and our DTTO, respectively.
As can be seen, the objects in existing VOT datasets consistently maintain their categories throughout the sequences. In contrast, examples from our DTTO dataset demonstrate objects undergoing category changes, specifically transitioning from a fighter jet to a car and from a seed to a young sprout. 
These examples suggest that category changes in tracking transforming objects introduce complex transformations wherein objects undergo more significant changes in appearance, shape, and context compared to generic object tracking. The diverse nature of category changes requires algorithms to adapt to varying object appearances and environmental conditions, further complicating the tracking process. Additionally, the temporal consistency of object tracking becomes crucial in the presence of category changes, as algorithms must accurately track objects across frames while accommodating the dynamic nature of transformations. Therefore, tracking transforming objects present a unique and challenging task for object tracking algorithms, requiring robust solutions capable of handling complex transformations and maintaining accurate tracking over time.
By presenting DTTO, a diverse dataset encompassing a wide array of transformation processes, environments, and object categories, we aim to equip researchers and practitioners with the necessary resources to tackle the challenges inherent in tracking transforming objects. Moreover, we seek to elevate awareness and interest in advanced tracking tasks that transcend the confines of traditional object tracking, thereby fostering innovation and progress in the field of visual tracking. Our contributions can be summarized as follows,

\begin{itemize}
% \item We introduce a critical yet under-explored challenge in State-Changing Object Tracking (SCOT).

% \item We are starting our first exploration of tracking transforming objects, which we believe will provide a new perspective on visual object tracking and encourage the development of more advanced applications.

% \item We define a novel task of state-changing object tracking, which serves as the foundation for a range of applications requiring information about object states, yet is often overlooked in traditional Visual Object Tracking (VOT). This task may also draw attention to more informative object tracking methods that extend beyond conventional objects.

\item We make the first exploration of tracking transforming objects, aiming to uncover valuable insights that have the potential to innovate the way we perceive and approach object tracking tasks. This exploration is poised to offer a fresh perspective on the intricacies and challenges inherent in tracking objects undergoing transformations.

\item We presented the DTTO, the first benchmark dedicated to tracking transforming objects, presenting a unique and challenging visual task.

\item We conduct comprehensive evaluations of DTTO by using 20 representative trackers and analyzing their performance using diverse evaluation metrics, providing baselines for future comparison. 
% The results show considerable room for enhancing existing SOTA trackers, emphasizing the imperative of innovation in tracking objects under transformation.

\end{itemize}

\section{Related Work}
\subsection{Datasets for Generic Visual Object Tracking}

Early-stage tracking benchmarks are typically modest in size to ensure impartial assessment and comparison of various trackers.
Specifically, OTB-2013  \cite{wu2013online}  initiated the development of object tracking benchmarks by presenting 50 videos. This benchmark was subsequently expanded in OTB-2015 \cite{wu2015otb} with the addition of more videos, resulting in a total of 100.
VOT \cite{kristan2016novel} presents a fully-annotated dataset featuring per-frame annotations and multiple visual attributes, offering a range of challenges for comparing trackers across various aspects.
% NfS \cite{kiani2017need} presents 100 sequences with a high frame rate for the evaluation of tracking performance.
Several large-scale benchmarks (i.e., GOT-10k \cite{huang2019got}, LaSOT \cite{fan2019lasot}, and TrackingNet \cite{muller2018trackingnet}) have been introduced in recent years to address the inadequacy of early datasets in meeting the requirements of deep tracking training videos.
LaSOT consists of 1,400 long-term videos densely well-annotated, and it was subsequently expanded in \cite{fan2021lasot} with the inclusion of additional video sequences. 
% It is noteworthy that it provides both bounding boxes and language annotations, providing support for visual tracking as well as visual-language tracking.
TrackingNet introduces a large-scale dataset comprising approximately 30k videos for training deep tracking models.
Similarly, Got-10k provides around 10,000 video sequences across 563 classes, making it a sizable benchmark as well.
Please refer to Table \ref{dataset-compare} for a more information of these existing datasets.

Despite the availability of the above benchmarks, their primary focus is on tracking common objects, neglecting the challenges presented by transforming objects with significant changes of appearances, color, shape, and even object category, which are notably more challenging to track.
However, tracking transforming objects holds significant importance across various fields due to the dynamic nature of many real-world scenarios, including autonomous systems, human-computer interaction, and security applications.
In this work, we initiate an exploration of tracking transforming objects, aspiring to inspire and catalyze the development of more advanced methodologies and algorithms for  this challenging task.
Table \ref{dataset-compare} shows the comparison of DTTO with existing VOT datasets.

\begin{table*}[t]
\centering
\scriptsize
\setlength\tabcolsep{2.5pt}
\caption{Comparison of DTTO with existing generic tracking datasets.``Tra." and ``Eva." denote training and evaluation, respectively, while CO and TO represent common objects and transforming objects, respectively.}
\begin{tabular}{ccccccccc}
\toprule[1pt] 
\textbf{Benchmark} & \textbf{Year} & \textbf{Classes} & \textbf{Video} & \textbf{\begin{tabular}[c]{@{}c@{}}Mean\\ Frames\end{tabular}} & \textbf{\begin{tabular}[c]{@{}c@{}}Total\\ Frames\end{tabular}} & \textbf{\begin{tabular}[c]{@{}c@{}}Total\\ Duration\end{tabular}} & \textbf{\begin{tabular}[c]{@{}c@{}}Dataset\\ Goal\end{tabular}} & \textbf{Focus}                                                   \\ \hline \hline
OTB-2013\cite{wu2013online}           & 2013          & 10               & 50             & 578                                                            & 29K                                                             & 16.4 min                                                          & Eva.                                                            & CO          \\
OTB-2015\cite{wu2015otb}           & 2015          & 16               & 100            & 590                                                            & 59K                                                             & 32.8 min                                                          & Eva.                                                            & CO          \\
TC-128 \cite{liang2015encoding}             & 2015          & 27               & 128            & 429                                                            & 55K                                                             & 30.7 min                                                          & Eva.                                                            & CO          \\
NUS-PRO\cite{li2015nus}\            & 2016          & 17               & 365            & 371                                                            & 135K                                                            & 75.2 min                                                          & Eva.                                                            & CO          \\
UAV123\cite{mueller2016benchmark}             & 2016          & 9                & 123            & 915                                                            & 113K                                                            & 62.5 min                                                          & Eva.                                                            & CO          \\
UAV20L\cite{mueller2016benchmark}             & 2016          & 5                & 20             & 2,934                                                          & 59K                                                             & 62.6 min                                                          & Eva.                                                            & CO          \\
NfS\cite{kiani2017need}                & 2017          & 24               & 100            & 3,830                                                          & 383K                                                            & 26.6 min                                                          & Eva.                                                            & CO          \\
VOT-2017\cite{kristan2016novel}           & 2017          & 24               & 60             & 356                                                            & 21K                                                            & 11.9 min                                                          & Eva.                                                            & CO          \\
TrackingNet\cite{muller2018trackingnet}        & 2018          & 27               & 30,643         & 471                                                            & 14.43M                                                          & 140.0 hours                                                       & Tra./Eva.                                                       & CO          \\
LaSOT\cite{fan2019lasot}              & 2019          & 70               & 1400           & 2,053                                                          & 3.52M                                                           & 32.5 hours                                                        & Tra./Eva.                                                       & CO          \\
Got-10k\cite{huang2019got}            & 2021          & 563              & 9,935          & 149                                                            & 1.45M                                                           & 40.0 hours                                                         & Tra./Eva.                                                       & CO          \\
\rowcolor[HTML]{C0C0C0} 
\textbf{DTTO(Ours)}               & \textbf{2024}          & \textbf{11}               & \textbf{100}            & \textbf{93}                                                             & \textbf{9.3K}                                                            & \textbf{5.3 min}                                                           & \textbf{Eva.}                                                            & \textbf{TO}
\\ \toprule[1pt] 
\end{tabular}
\label{dataset-compare}
\end{table*}

\subsection{Visual Tracking Algorithms}

In recent years, deep neural networks have played a major role in the significant progress made in visual object tracking . 
The Siamese tracking framework \cite{bertinetto2016fully} has garnered significant attention in the tracking community for its commendable balance between accuracy and efficiency, resulting in the development of numerous extensions aimed at enhancing its performance \cite{fan2019siamese,fu2021stmtrack,guo2020siamcar,zhang2019deeper}.
More recently, the Vision Transformer model \cite{dosovitskiy2020image,vaswani2017attention} has been introduced into the field of visual object tracking due to its exceptional capability in context modeling within images.
For example, Chen et al. \cite{chen2021transformer} proposed the integration of Transformer into Convolutional Neural Network (CNN) architecture, yielding promising results.
Yan et al. \cite{yan2021learning} proposes a spatio-temporal Transformer network aimed at achieving better tracking performance.
To achieve a better balance between performance and inference speed, Ye et al. \cite{ye2022joint} propose a one-stream architecture for Transformer tracking that effectively integrates feature learning and relation modeling. 
Additionally, recent studies have introduced various transformer-based trackers \cite{wu2023dropmae,cai2023robust,kou2024zoomtrack,gao2023generalized}, further advancing the field of visual tracking. In this study, we undertake a thorough evaluation of the performance of 20 state-of-the-art trackers in the challenging task of tracking transforming objects, leveraging the newly introduced DTTO dataset, through which we aim to provide insights into the effectiveness and robustness of these trackers in handling the complexities posed by transforming objects across various scenes and scenarios captured in DTTO.

\section{The Proposed DTTO Dataset}

Our objective is to develop a dedicated \textbf{D}ataset for \textbf{T}racking \textbf{T}ransforming \textbf{O}bjects (DTTO).
The primary steps in developing the DTTO involve selecting representative videos spanning diverse object categories and transformation processes, followed by manual annotations for each video sequence, as detailed later.

% \begin{figure}[h]
%   \centering
%   \subfigure{\includegraphics[width=0.49\textwidth]{LaTeX2e_Proceedings_Templates/figs/object-types.png}}
%   \hfill
%   \subfigure{\includegraphics[width=0.49\textwidth]{LaTeX2e_Proceedings_Templates/figs/action-types.png}}
%   \caption{11}
%   \label{statics}
% \end{figure}

% \begin{figure}[h]
%   \centering
%   \subfigure{\includegraphics[width=0.49\textwidth]{LaTeX2e_Proceedings_Templates/figs/action-dist.png}}
%   \hfill
%   \subfigure{\includegraphics[width=0.49\textwidth]{LaTeX2e_Proceedings_Templates/figs/correlation-dataset.png}}
%   \caption{Statistics from DTTO: on the left, the distribution of transformations, and on the right, the co-occurrence statistics between the transformations and object categories.}
%   \label{statics}
% \end{figure}

\begin{figure*}[h]
	\centering
\includegraphics[width=1.0\textwidth]{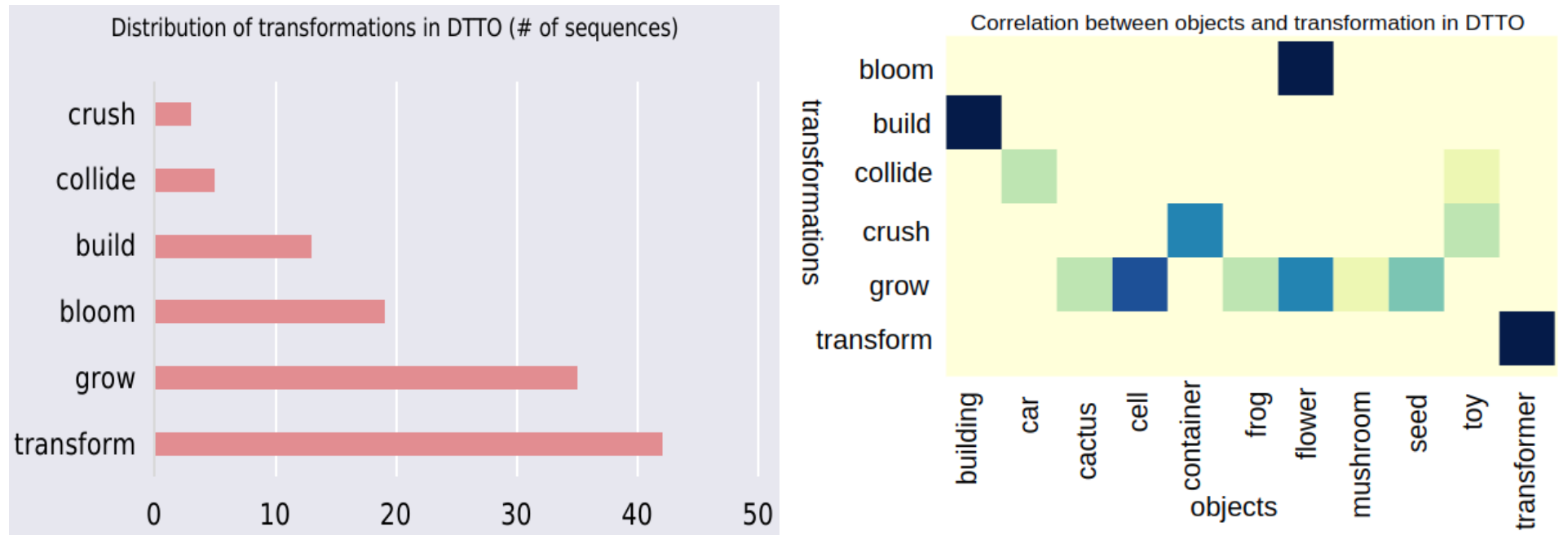}
   \caption{Statistics of DTTO: on the left, the distribution of transformations, and on the right, the co-occurrence statistics between the transformations and object categories.}
    \label{statics}
\end{figure*}

\subsection{Data Collection}

Our benchmark aims to cover common object classes that undergo transformations during tracking.
To achieve this goal, we select 11 categories, including `building', `car', `cell', `cactus', `container', `flower', `frog', `mushroom', `seed', `toy', and `transformer', to develop the DTTO.
Among them, we present six types of transformations: `bloom', 
`build', `collide', `crush', `grow', and `transform'.
After identifying the object categories and transformation types, we collected a variety of video sequences from the Internet, as it serves as a primary source for many tracking benchmarks (such as LaSOT \cite{fan2019lasot}, GOT-10k \cite{huang2019got}, TrackingNet \cite{muller2018trackingnet}, etc.).
At first, we collected more than 150 sequences, totaling over 30K frames.
% After carefully inspecting the availability of each video, we selected 100 sequences and filtered out irrelevant frames, resulting in an average video length of 93 frames per sequence. These sequences present objects that have undergone a notable transformation process.
While all the previously selected sequences exhibit transforming objects, not all lead to a significant transformation. Therefore, we eliminate irrelevant sequences and redundant frames to ensure that the objects presented in the remaining sequences undergo a substantial transformation process.
Ultimately,  the DTTO dataset comprises 100 video sequences, totaling 9,297 frames, with an average video length of approximately 93 frames.
% Fig. \ref{statics}a and \ref{statics}b present histogram data regarding the categories of objects and transformation within the entire dataset, respectively.
Fig. \ref{statics} presents the histogram depicting the distribution of transformation types across video sequences (left) and co-occurrence statistics between transformation types and object categories (right), respectively. As evident, the 'transform', 'grow', and 'bloom' transformations are the most common, collectively constituting over 75\% of all sequences. It's noteworthy that these transformations involve changes in object categories, underscoring the conceptual complexities inherent in our dataset.
%ore than  As shown in  mthe figure, the `transform' and `grow' actions are the most prevalent.
In addition, there is significant entropy in the correlation statistics, highlighting the diversity of our dataset.

\begin{figure*}[h]
	\centering
\includegraphics[width=1.0\textwidth]{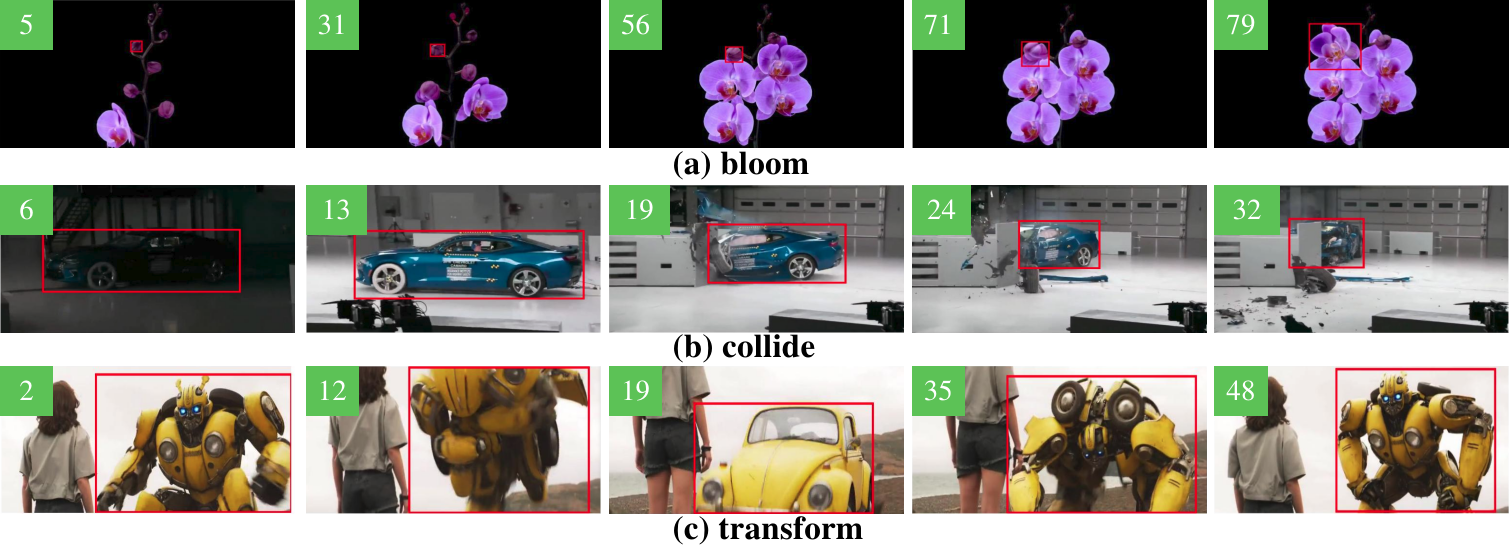}
        \caption{Visual representation of annotation examples for three types (i.e., bloom, collide, and transform) within the proposed DTTO.}
 \label{anno_figs}
\end{figure*}

% \begin{figure}[htbp]
%     \centering
%     \begin{minipage}[b]{0.95\textwidth}
%         \centering
%         \includegraphics[width=\textwidth]{LaTeX2e_Proceedings_Templates/figs/sample-bloom.png}
%         \subcaption{(a) bloom}
%     \end{minipage}
%     \hfill
%     \begin{minipage}[b]{0.95\textwidth}
%         \centering
%         \includegraphics[width=\textwidth]{LaTeX2e_Proceedings_Templates/figs/sample-collide.png}
%         \subcaption{(b) collide}
%     \end{minipage}
%         \begin{minipage}[b]{0.95\textwidth}
%         \centering
%         \includegraphics[width=\textwidth]{LaTeX2e_Proceedings_Templates/figs/sample-transform.png}
%         \subcaption{(c) transform}
%     \end{minipage}
%         \caption{Visual representation of annotation examples for three types (i.e., bloom, collide, and transform) within the proposed DTTO.}
%  \label{anno_figs}
% \end{figure}

\subsection{Annotation}
% Following the annotating principles \cite{lasot1 ,lasot2}, starting with the first target item, the expert (i.e., a student engaged in object tracking) carefully draws the narrowest (axis-aligned) bounding box around any visible part of the target in each frame. It is noteworthy to highlight that items that are out of visual range are not included in this collection, and we do not assign corresponding attribute labels to objects with attributes such as occlusion and motion blur.

During annotating data, we adhere the principle provided in the annotation guide \cite{fan2019lasot,fan2021lasot}. For each frame in the sequence, the annotator generates an axis-aligned bounding box to tightly encompass any visible part of the object, beginning from an initial object; otherwise, an absence label is assigned to the frame, indicating either being out-of-view or fully occluded.
% It's important to highlight that, as our focus is on object tracking under transformation, this dataset excludes objects that are either out of view (OV) or full occlusion (FOC). Consequently, throughout the annotation process, no frame will be assigned with an absence label.
Based on the above principle, we finish the annotation in three steps: manual labeling, visual inspection, and refinement.
In the first step, the expert (i.e., a student engaged in object tracking) begins labeling the target in all frames of the video.
It is important to note that, during the annotation process, each video sequence is labeled by the same annotator to ensure consistency.
Given that annotation errors or inconsistencies are inevitable in the initial stage, a visual inspection is conducted by the verification team in the second step to validate the annotations.  If the validation team deems that there are still concerns with the annotations, they will be sent back to the original annotator for refinement in the third step.
During the annotation process, the second and third steps are repeated multiple times  to establish a benchmark with annotations of superior quality.
Fig. \ref{anno_figs} displays some representative examples of bbox annotations from DTTO.

\begin{wrapfigure}[11]{r}{0.5\textwidth}
  \begin{minipage}{\linewidth}
    \centering
        \vspace{-24pt}
    \includegraphics[width=0.9\linewidth]{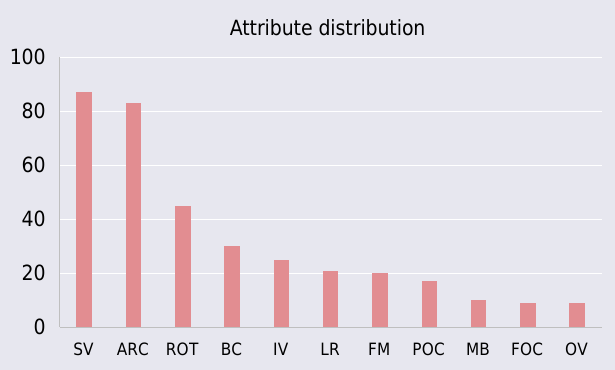} % 替换为你的图片
  \end{minipage}
      \caption{Distribution of sequences per attribute.}
   \label{attr-dist}
\end{wrapfigure}

% \subsection{Dataset Statistics}
% % Our SCVOT dataset comprises 100 video clips showcasing state changes, with each clip averaging a duration of 7.4 seconds. These videos cover 5 action categories and 10 object categories, resulting in a total of 271 valid combinations.

% Fig. \ref{statics} displays the distribution of statistical data within the DTTO.
% Specifically, Fig. \ref{statics}a and \ref{statics}b present histogram data regarding the categories of transformation and objects within the entire dataset, respectively.
% It is worth noticing that the distribution of data among various categories of transformations and objects follows a long-tail distribution, which is a phenomenon commonly observed in the real world.
% The Transformers object and its transformation action are the most prevalent, constituting approximately half of the action and object categories.
% Overall, there is significant entropy in the correlation statistics, highlighting the diversity of our dataset.

\subsection{Attributes}

In order to conduct a thorough analysis of tracking methods, we provide eleven attributes commonly found in visual tracking tasks, including (1) Scale Variation (SV), which is
assigned when the ratio of the bounding box is outside the range [0.5, 2], (2) Aspect Ratio Change (ARC), which is assigned when the ratio of the bounding box aspect ratio is outside the range [0.5, 2], (3) Rotation (ROT), (4) Background Clutter (BC), (5) Illumination Variation (IV), (6) Low Resolution (LR), which is assigned when the target area is smaller than 900 pixels, (7) Fast Motion (FM), which is assigned when the target center moves by at least 50\% of its size in the last frame, (8) Partial Occlusion (POC), (9) Motion Blur (MB), (10) Full Occlusion (FOC), and (11) Out-of-View (OV).
% Our focus in this work lies in tracking under transformations, with deformation being the fundamental attribute in our benchmark. 
Each sequence is annotated with an 11-dimensional vector indicating the presence or absence of each attribute.
It is noteworthy that we do not treat deformation as an isolated attribute, but rather consider it inherent in all video sequences by default.
Fig.\ref{attr-dist} presents the distribution of attributes for DTTO.
As evident, the three most frequent challenges are Scale Variation, Aspect Ratio Change, and Rotation, which can be attributed to the transformation inherent in each video.

\section{Experiments}

\subsection{Evaluation Metrics}
\label{exper:metrics}

Following established benchmarks such as \cite{fan2019lasot,muller2018trackingnet,wu2013online}, we employ one-pass evaluation (OPE) for assessing and comparing trackers. This evaluation method utilizes two standard metrics: precision (PRC) and success rate (SUC).
More specifically, PRC calculates the pixel-wise distance between the center positions of tracking results and ground truth, with trackers being ranked based on their PRC using a predetermined threshold, for instance, 20 pixels.
% Taking into account the influence of various video resolutions, NPRE is calculated by normalizing PRC with respect to the target region to mitigate this issue. 
However, SUC assesses the Intersection over Union (IoU) between tracking results and ground truth boxes, determining the percentage of frames where the IoU exceeds a specified threshold (e.g., 0.5).

\begin{wraptable}[17]{r}{0.5\linewidth}
\centering
\vspace{-24pt}
\scriptsize
% \caption{Summary of Algorithms:``CNN", ``CNN-T", and ``Trans." denote CNN-based, CNN-Transformer-based, and Transformer-based trackers, respectively.} 
\caption{Summary of tracking models.} 
\begin{tabular}{cccc}
\toprule[1pt] 
Tracker   & Source     & Backbone   & Type   \\ \hline \hline
Ocean\cite{zhang2020ocean}     & CVPR 20    & ResNet-50  & CNN    \\
SiamCAR\cite{guo2020siamcar}   & CVPR 20    & ResNet-50  & CNN    \\
PrDiMP\cite{danelljan2020probabilistic}    & CVPR 20    & ResNet-50  & CNN    \\
TransT\cite{chen2021transformer}    & CVPR 21    & ResNet-50  & CNN-T  \\
AutoMatch\cite{zhang2021learn} & ICCV  21   & ResNet-50  & CNN    \\
STARK\cite{yan2021learning}     & ICCV  21   & ResNet-101 & CNN-T  \\
RTS\cite{paul2022robust}       & ECCV 22    & ResNet-50  & CNN    \\
ToMP\cite{mayer2022transforming}      & CVPR 22    & ResNet-101 & CNN-T  \\
MixFormer\cite{cui2022mixformer} & CVPR 22    & CVT21    & Trans. \\
OSTrack\cite{ye2022joint}   & ECCV 22    & ViT-Base   & Trans. \\
SeqTrack\cite{chen2023seqtrack}  & CVPR 23    & ViT-Base   & Trans. \\
DropTrack\cite{wu2023dropmae} & CVPR 23    & ViT-Base   & Trans. \\
GRM\cite{gao2023generalized}       & CVPR 23    & ViT-Base   & Trans. \\
HiT\cite{kang2023exploring}       & ICCV 23    & LeViT-384  & Trans. \\
ROMTrack\cite{cai2023robust}  & ICCV 23    & ViT-Base   & Trans. \\
ZoomTrack\cite{kou2024zoomtrack} & NeurIPS 23 & ViT-Base   & Trans. \\
LightFC\cite{li2024lightweight}   & KBS 24     & ViT-Tiny   & Trans. \\
DCPT\cite{zhu2024dcpt}      & ICRA 24    & ViT-Base   & Trans. \\
EVPTrack\cite{shi2024explicit}  & AAAI 24    & HiViT-Base & Trans. \\
AQATrack\cite{xie2024autoregressive}  & CVPR 24    & HiViT-Base & Trans. \\ \toprule[1pt] 
    \end{tabular}
    \label{tracker-list}
\end{wraptable}

\subsection{Evaluated Trackers}

We conducted a comprehensive evaluation of 20 representative trackers, spanning various time periods, to evaluate the performance of current trackers and establish baselines for future comparisons on DTTO. These trackers were categorized into three different groups for analysis purposes:
\\ \textbf{(i) CNN-based trackers} that achieve object tracking solely through CNN architecture, including Ocean \cite{zhang2020ocean}, SiamCAR \cite{guo2020siamcar}, PrDiMP \cite{danelljan2020probabilistic}, AutoMatch \cite{zhang2021learn}, and RTS \cite{paul2022robust}.
\\ \textbf{(ii) CNN-Transformer-based trackers} that implement visual tracking through a hybrid backbone architecture combining CNNs and Transformers consisting of TransT \cite{chen2021transformer}, STARK \cite{yan2021learning}, and ToMP \cite{mayer2022transforming}.
\\ \textbf{(iii) Transformer-based trackers} that track objects by leveraging a pure Transformer backbone architecture. The visual object tracking models within this category include MixFormer \cite{cui2022mixformer}, OSTrack \cite{ye2022joint}, SeqTrack \cite{chen2023seqtrack}, DropTrack \cite{wu2023dropmae}, GRM \cite{gao2023generalized}, HiT \cite{kang2023exploring}, ROMTrack \cite{cai2023robust}, ZoomTrack \cite{kou2024zoomtrack}, LightFC  \cite{li2024lightweight}, DCPT \cite{zhu2024dcpt}, EVPTrack \cite{shi2024explicit}, and AQATrack \cite{xie2024autoregressive}.

The trackers previously mentioned are evaluated in their original implementation without any modifications. A summary of these tracking algorithms is provided in Table \ref{tracker-list}.
As shown in the table, ``CNN", ``CNN-T", and ``Trans." denote CNN-based, CNN-Transformer-based, and Transformer-based trackers, respectively.

\begin{figure}[t]
    \centering
    \begin{minipage}[b]{0.48\textwidth}
        \centering
        \includegraphics[width=\textwidth]{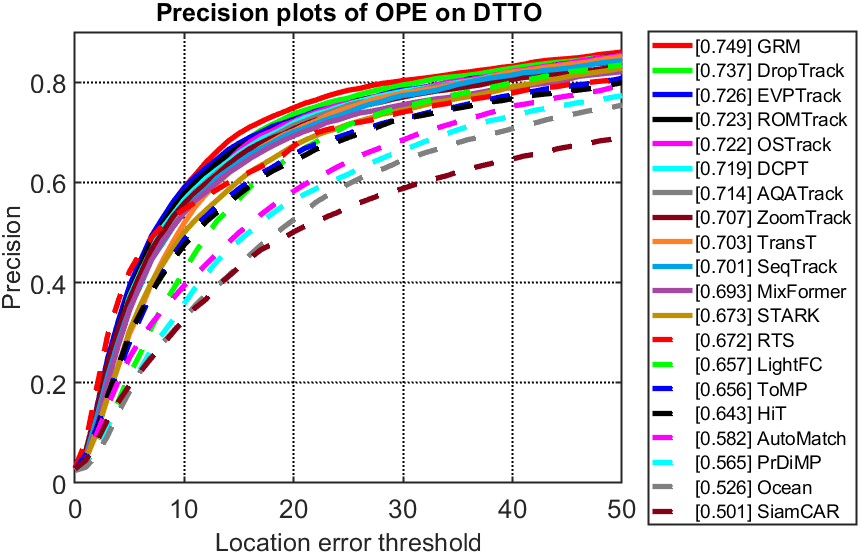}
    \end{minipage}
    \begin{minipage}[b]{0.48\textwidth}
        \centering
        \includegraphics[width=\textwidth]{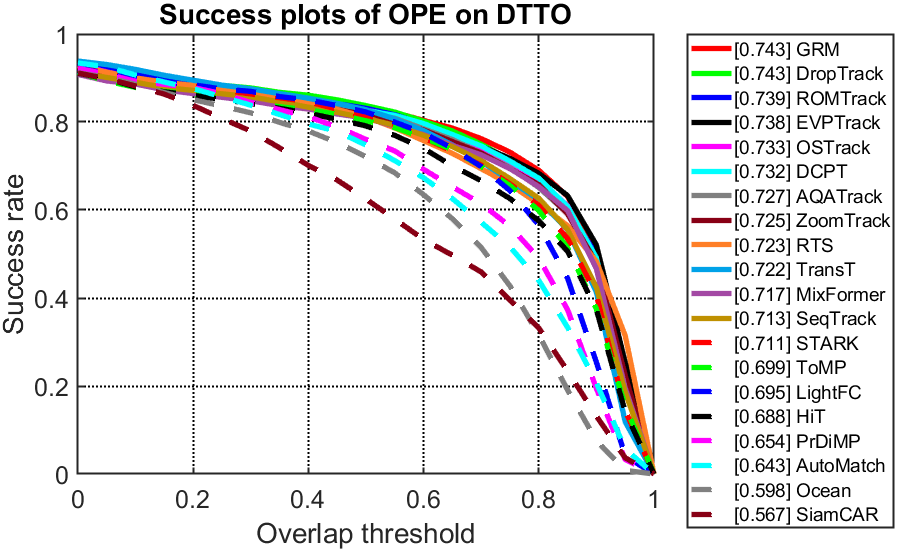}
    \end{minipage}
        \caption{The overall performance of 20 SOTA trackers are evaluated on DTTO. Precision and success rate, as defined by one-pass evaluation (OPE)\cite{wu2013online}, are employed for evaluation.}
 \label{prec_succ_figs}
\end{figure}

\subsection{Evaluation Results}

\textbf{Overall Performance.} We employ 20 state-of-the-art trackers, as summarized in the previous subsection, to conduct a thorough evaluation on DTTO. It's worthy of note that all trackers are utilized without any modifications to ensure a fair and truthful comparison. The precision and success plots of OPE are depicted in Fig. \ref{prec_succ_figs}, providing insights into the performance of each tracker across various attributes and scenarios present in DTTO.
As observed, the top three trackers in terms of PRC are GRM, DropTrack, and EVPTrack, achieving PRC scores of 74.9\%, 73.7\%, and 72.6\%, respectively. Regarding SUC score, GRM and DropTrack exhibit optimal performance, both achieving a SUC score of 74.3\%. Following closely, ROMTrack and EVPTrack secure the second and third-best SUC scores of 73.9\% and 73.8\%, respectively. Notably, all these leading trackers are built using the vision Transformer architecture, demonstrating its power in feature learning for tracking.
Interestingly, despite not leveraging Transformer architecture, RTS achieves notable results, with PRC and SUC scores of 67.2\% and 72.3\%, respectively. Remarkably, its performance surpasses that of certain Transformer-based trackers, such as LightFC, which attained PRC and SUC scores of 65.7\% and 69.5\%, and HiT, with scores of 64.3\% for PRC and 68.8\% for SUC. We attribute this to RTS's utilization of tracking-by-segmentation, which proves beneficial for tracking objects undergoing transformations. By segmenting objects into coherent regions, RTS can maintain tracking accuracy even amidst significant transformations in appearance and shape. This highlights the importance of considering tailored tracking strategies beyond Transformer-based approaches, especially in scenarios involving dynamic object transformations. 

\begin{figure*}[t]
	\centering
    \subfigure{
		\begin{minipage}[t]{0.3\textwidth}
			\includegraphics[width=1\textwidth,height=0.95\textwidth]{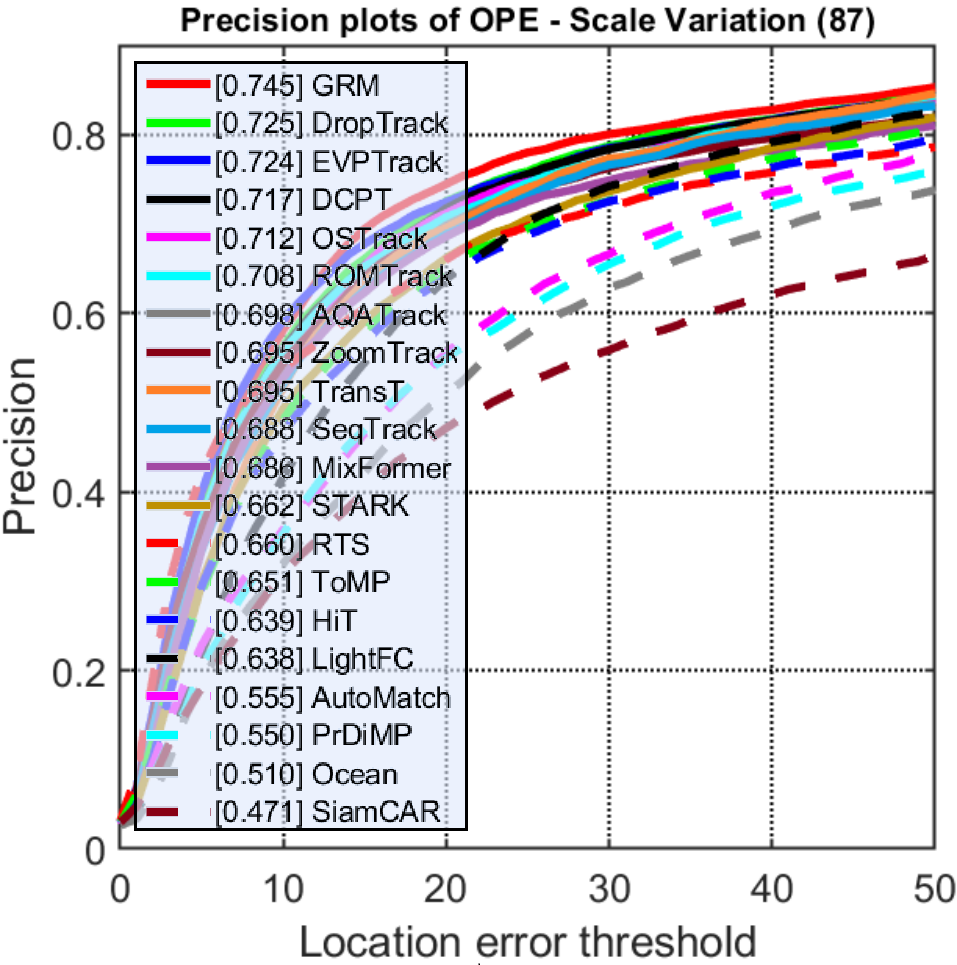}\hspace{0in}
   			\includegraphics[width=1\textwidth,height=0.95\textwidth]{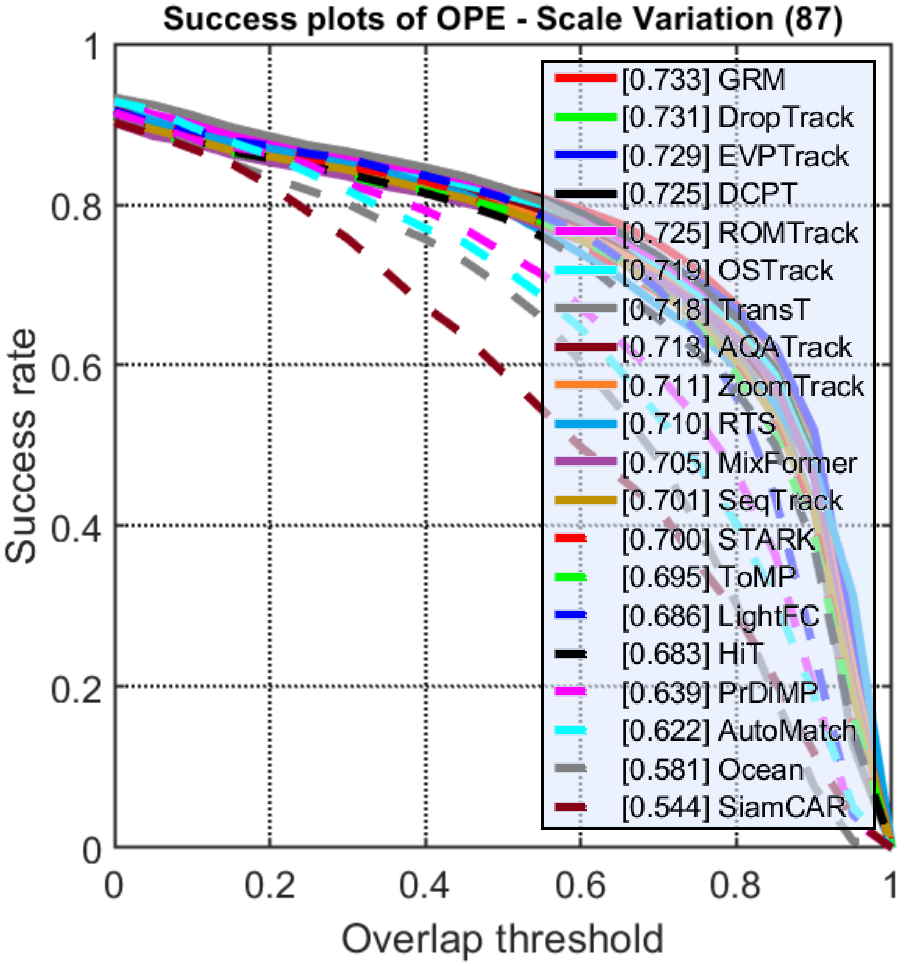}\hspace{0in}
	   \end{minipage}}
	\subfigure{
		\begin{minipage}[t]{0.3\textwidth}
			\includegraphics[width=1\textwidth,height=0.95\textwidth]{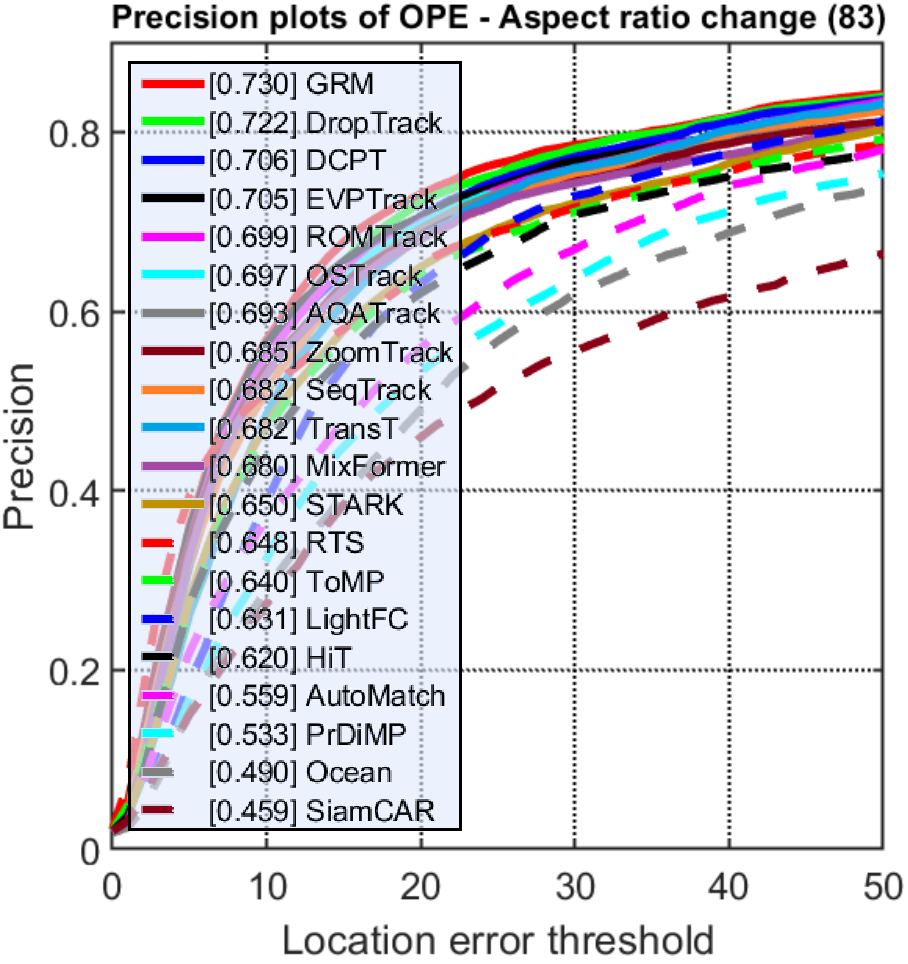}\hspace{0in}
            \includegraphics[width=1\textwidth,height=0.95\textwidth]{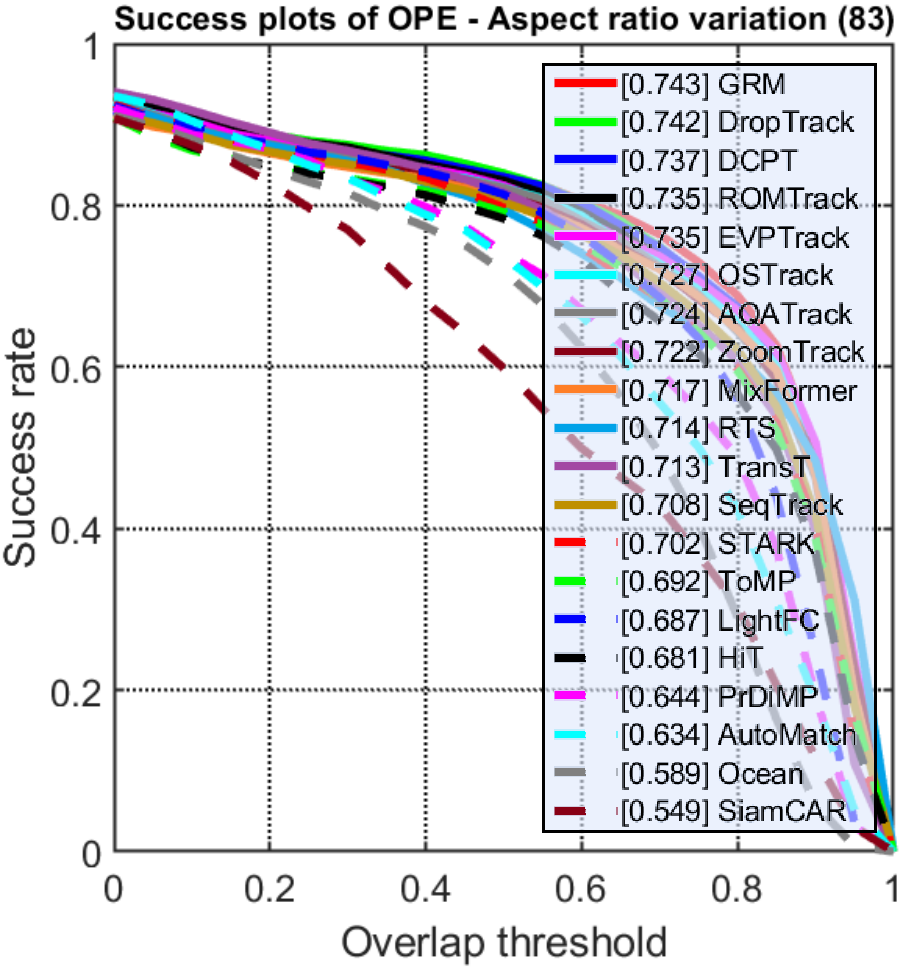}\hspace{0in}
            \end{minipage}}
	\subfigure{
		\begin{minipage}[t]{0.3\textwidth}
			\includegraphics[width=1\textwidth,height=0.95\textwidth]{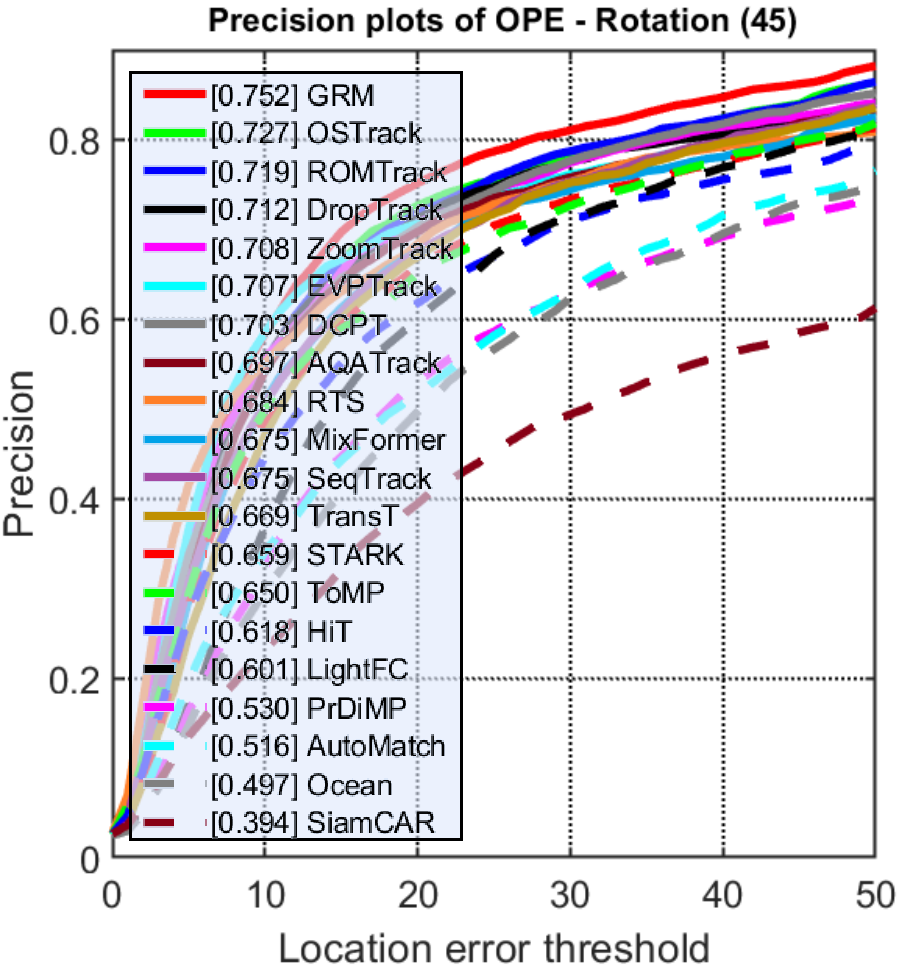}\hspace{0in}
              \includegraphics[width=1\textwidth,height=0.95\textwidth]{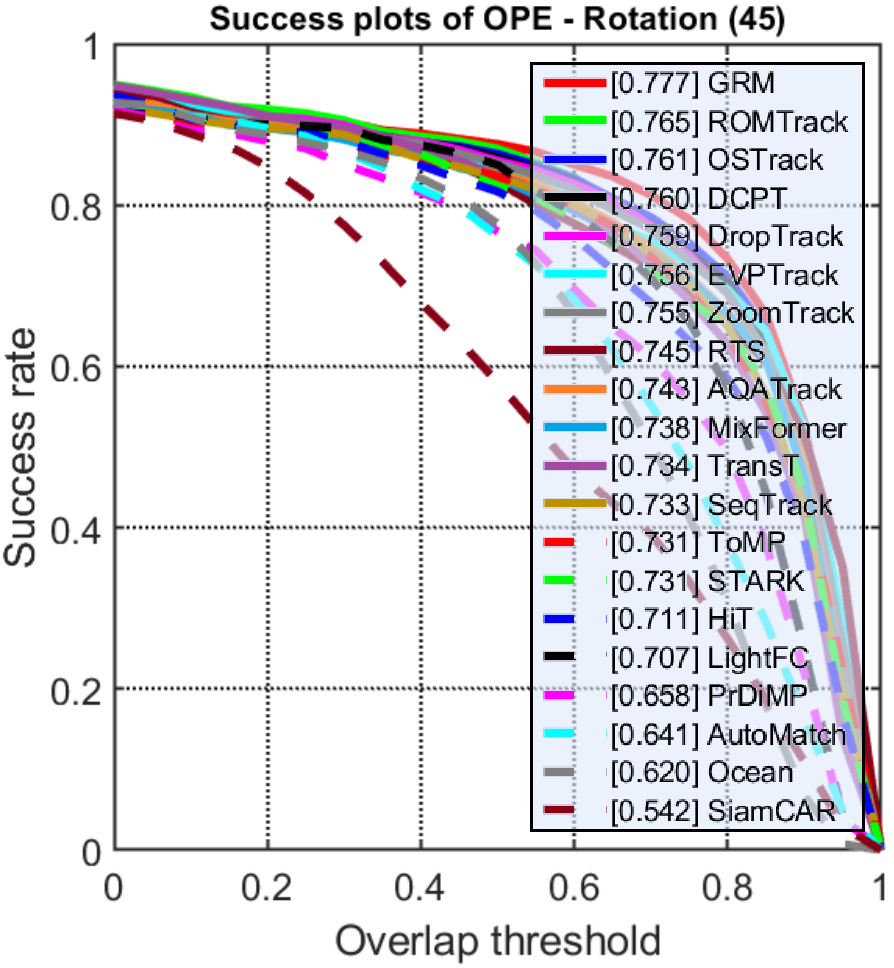}\hspace{0in}
			
	   \end{minipage}}

    \caption{Comparison based on attributes across the three most common attributes: scale variation (SV) aspect ratio change (ARC), and rotation (ROT).}
 \label{attr-figs}
\end{figure*}

\textbf{Attribute-based performance.}
To gain a better analysis of the differences in performance across various trackers, we evaluated their performance across eleven attributes. 
Due to page limitations, Fig. \ref{attr-figs} displays the attribute-based evaluation results for the three most prevalent attributes: SV, ARC, and ROT.
% As illustrated in Fig. \ref{attr-figs} (a), GRM and DropTrack exhibit the two highest precision of 0.730/0.745 and 0.722/0.725, respectively, making them the top two trackers in ASC and SV. 
As illustrated in Fig. \ref{attr-figs} (a),  GRM demonstrates the highest precision, i.e., 74.5\%, 73.0\%, and 75.2\%, across all three attribute subsets, with DropTrack and OSTrack securing the second position. Specifically, DropTrack ranks second in ARC and SV with precisions of 72.2\% and 72.5\%, while OSTrack ranks second in ROT with 72.7\% precision.
% GRM and DropTrack exhibit the two highest performance of (0.730, 0.745) and (0.722/0.725 in ARC and SV, respectively, which is consistent with their performance in overall evaluation.
% Similarly, GRM and OSTrack showcase the optimal two results in ROT, achieving precision of 0.752 and 0.727, respectively.
In terms of success rate, as shown in Fig. \ref{attr-figs} (b), GRM and DropTrack occupy the top two positions in ASC and SV. Specifically, their SUCs are 74.3\% and 74.2\% for ASC, and 73.3\% and 73.1\% for SV, respectively. Additionally, GRM and ROMTrack achieve the best two results in ROT with 77.7\% and 76.5\%, respectively. These results suggest that GRM and DropTrack are basically the  top-performing trackers on these attribute subsets.

\begin{table}[h]
\scriptsize
\centering
\setlength\tabcolsep{3.0pt}
\caption{Illustration of the performance of per transformation type in terms of precision (PRC) and success rate (SUC) on DTTO, shown in the form of (PRC, SUC). The optimal results are highlighted in bold.}
\begin{tabular}{ccccccc}
\toprule[1pt] 
          & bloom       & build        & collide     & crush       & grow        & transform \\ \hline \hline
GRM\cite{gao2023generalized}       & (82.7,80.3) & (81.3,82.0) & (57.6,76.5) & (47.3,52.9) & (\textbf{71.3,66.5}) & (74.6,79.6)  \\
DropTrack\cite{wu2023dropmae} & (\textbf{87.2,83.3}) & (76.8,80.8) & (62.4,83.8) & (\textbf{52.8,56.3}) & (71.0,66.3) & (74.3,80.8)  \\
EVPTrack\cite{shi2024explicit}  & (84.7,83.7) & (76.6.80.6) & (59.6,82.7) & (50.1,55.8) & (66.8,70.1) & (73.8,79.9)  \\
ROMTrack\cite{cai2023robust}  & (83.0,80.5) & (77.4,82.4) & (\textbf{66.1,84.8}) & (47.6,53.8) & (67.4,65.5) & (73.9,80.7)  \\
OSTrack\cite{ye2022joint}   & (84.1,82.8) & (77.4,82.9) & (60.4,82.7) & (45.1,49.3) & (64.3,62.5) & (\textbf{75.2,80.9})  \\
DCPT\cite{zhu2024dcpt}      & (87.0,83.1) & (\textbf{83.2,83.0}) & (63.9,84.3) & (49.1,45.2) & (60.6,60.1) & (73.1,80.9)  \\
AQATrack\cite{xie2024autoregressive}  & (82.3,81.0) & (75.7,80.3) & (60.9,83.5) & (48.5,54.8) & (67.8,65.3) & (72.7,79.2)  \\
ZoomTrack\cite{kou2024zoomtrack} & (83.0,80.5) & (76.4,79.9) & (57.7,83.8) & (46.1,53.7) & (67.0,65.6) & (71.7,78.9) \\ \toprule[1pt] 
\end{tabular}
\label{actions-res}
\end{table}

% \begin{table}[h]
% \scriptsize
% \centering
% \setlength\tabcolsep{0.1pt}
% \caption{Illustration of the performance of various transformations in terms of precision (PRC) and AUC on DTTO. The optimal results are highlighted in bold.}
% \begin{tabular}{ccccccc}
% \toprule[1pt] 
%           & bloom       & build        & collide     & crush       & grow        & transform \\ \hline \hline
% GRM\cite{gao2023generalized}       & (0.827,0.803) & (0.813,0.820) & (0.576,0.765) & (0.473,0.529) & (\textbf{0.713,0.665}) & (0.746,0.796)  \\
% DropTrack\cite{wu2023dropmae} & (\textbf{0.872,0.833}) & (0.768,0.808) & (0.624,0.838) & (\textbf{0.528,0.563}) & (0.710,0.663) & (0.743,0.808)  \\
% EVPTrack\cite{shi2024explicit}  & (0.847,0.837) & (0.766,0.806) & (0.596,0.827) & (0.501,0.558) & (0.668,0.701) & (0.738,0.799)  \\
% ROMTrack\cite{cai2023robust}  & (0.830,0.805) & (0.774,0.824) & (\textbf{0.661,0.848}) & (0.476,0.538) & (0.674,0.655) & (0.739,0.807)  \\
% OSTrack\cite{ye2022joint}   & (0.841,0.828) & (0.774,0.829) & (0.604,0.827) & (0.451,0.493) & (0.643,0.625) & (\textbf{0.752,0.809})  \\
% DCPT\cite{zhu2024dcpt}      & (0.870,0.831) & (\textbf{0.832,0.830}) & (0.639,0.843) & (0.491,0.452) & (0.606,0.601) & (0.731,0.809)  \\
% AQATrack\cite{xie2024autoregressive}  & (82.3,81.0) & (75.7,80.3) & (60.9,83.5) & (48.5,54.8) & (67.8,65.3) & (72.7,79.2)  \\
% ZoomTrack\cite{kou2024zoomtrack} & (83.0,80.5) & (76.4,79.9) & (57.7,83.8) & (46.1,53.7) & (67.0,65.6) & (71.7,78.9) \\ \toprule[1pt] 
% \end{tabular}
% \label{actions-res}
% \end{table}

\textbf{Performance on per transformation type.} 
We employ the top eight trackers, i.e., GRM, DropTrack, EVPTrack, ROMTrack, OSTrack, DCPT, AQATrack, and ZoomTrack, to evaluate their performance on each transformation type, aiming to gain deeper analysis into the baseline trackers' performance in tracking transforming objects.
Table \ref{actions-res} presents the precision (PRC) and success rate (SUC) of these trackers on DTTO with the form of (PRC, SUC).
It is important to note that to emphasize the algorithms' superior performance in tracking `bloom' transformation, we intentionally separated the 'bloom' subset from the 'grow' category in this analysis.
It's evident that all the trackers exhibited exceptional performance during the 'bloom' transformation process, with all PRC and SUC surpassing 80\%.
However, their performance notably deteriorated during the 'crush' transformation, with PRC and SUC mostly falling below 50\%.
% This observation might stem from the flowers not changing significantly in amplitude across consecutive frames since they maintained a standard target size before and after blossoming. 
These results may be attributed to the smoother transformation process of flowers' blooming, with subtle changes in shape and minimal variations in appearance between frames, makes it relatively easier for trackers to maintain accurate tracking. The consistent background and relatively uniform texture of the petals further contribute to the tracking process's stability. In contrast, the sudden and drastic transformation associated with the crushing associated with colliding introduces significant challenges. The object undergoes a rapid change in size and shape, accompanied by a complex and irregular texture. This abrupt alteration disrupts the trackers' ability to maintain precise localization, leading to decreased tracking performance during the crushing transformation. 
% But the target changes from being a regular-sized item to one with a finer texture as it is crushed.
% In the former scenario, the inter-frame variation of the target, influenced by its standard size, remains relatively small, presenting a lesser challenge to the tracker. However, the complex deformations resulting from crushing cause substantial changes in the shape of targets between adjacent frames.
Notably, for most trackers, the difference between PRC and SUC in the `collide' type is more than 20\%. 
This is mainly attributed to PRC is more sensitive than SUC to the noticeable changes in the object's shape after the impact, as PRC reflects the accuracy of tracking results at fixed threshold while SUC aggregates overall performance which remains stable even under significant fluctuations.
These experimental results suggest that the performance of tracking transforming objects is closely tied to the complexity of the transformation processes those objects are experiencing.

% \begin{table}[h]
% \scriptsize
% \centering
% \setlength\tabcolsep{3.0pt} 
% \begin{tabular}{ccccccc}
% % \diagbox{Transformations}{Trackers}
%          & GRM                  & DropTrack            & EVPTrack    & ROMTrack    & OSTrack     & DCPT                 \\ \hline
% bloom       & (81.7,77.3)          & \textbf{(87.2,83.3)} & (84.7,83.7) & (83.0,80.5) & (84.1,82.8) & (87.0,83.1)          \\
% build        & (81.3,82.0)          & (76.8,80.8)          & (76.6.80.6) & (77.4,82.4) & (77.4,82.9) & \textbf{(83.2,83.0)} \\
% collide      & (57.6,76.5)          & (62.4,83.8)          & (59.6,82.7) & (66.1,84.8) & (60.4,82.7) & \textbf{(63.9,84.3)} \\
% crush       & (47.3,52.9)          & \textbf{(52.8,56.3)} & (50.1,55.8) & (47.6,53.8) & (45.1,49.3) & (49.1,45.2)          \\
% grow        & \textbf{(71.1,66.5)} & (71.0,66.3)          & (66.8,70.1) & (67.4,65.5) & (64.3,62.5) & (60.6,60.1)          \\
% transformate & \textbf{(74.6,79.8)} & (74.3,79.5)          & (73.8,79.9) & (73.9,80.7) & (75.2,80.9) & (73.1,80.9)         
% \end{tabular}
% \caption{Illustration of the performance of various transformations in terms of precision (PRC) and AUC on DTTO. The optimal results are highlighted in bold.}
% \label{actions-res}
% \end{table}

\begin{figure*}[t]
	\centering
\includegraphics[width=0.9\textwidth]{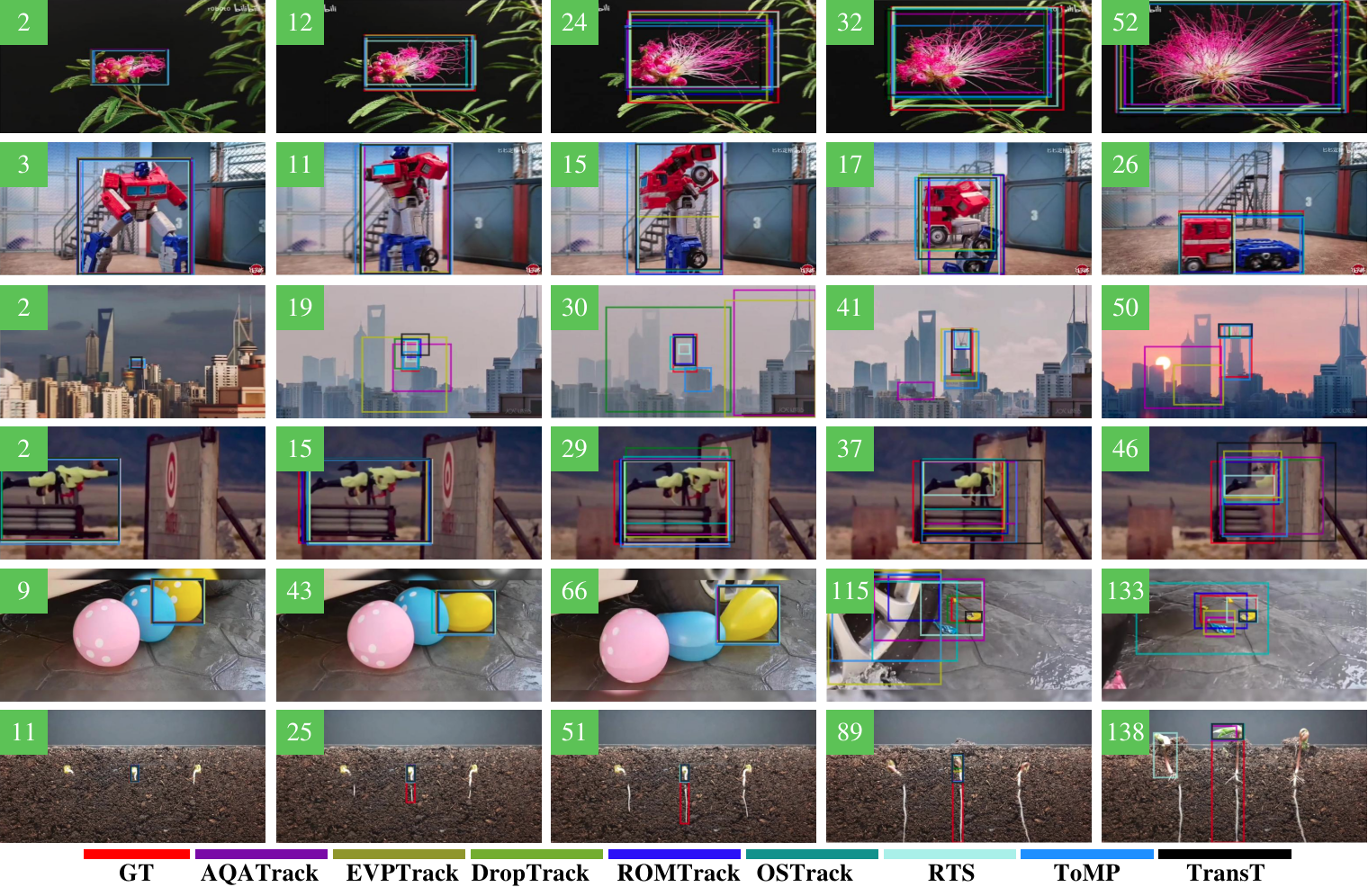}
        \caption{Qualitative evaluation conducted on six sequences from DTTO, i.e., flower3, transformer26, building7, toy1, container2, and seed5, arranged from top to bottom. Various methods' results are depicted using different colors, with `GT' representing the ground truth.}
 \label{figs-val}
\end{figure*}

\textbf{Qualitative evaluation.} 
In Fig. \ref{figs-val}, we use eight representative trackers (i.e., AQATrack, EVPTrack, DropTrack, ROMTrack, OSTrack, RTS, ToMP, and TransT) to illustrate qualitative tracking results for six different types of transformations in DTTO: bloom, transform, build, collide, crush, and grow.
As shown, for the first two rows, the trackers attain relatively satisfactory results compared to the rest examples, which can be attributed to the target remains stationary throughout the transformation process or is less affected by background clutter.
% Nevertheless, in the last four rows, due to irregular transformations and the impact of background clustering, the trackers fail to yield satisfactory results both before and after the target undergoes transformation.
Nevertheless, in the last four rows, due to background clustering and complex transformations wherein objects undergo more significant changes in appearance, shape, and context, the trackers fail to yield satisfactory tracking.
% The experimental results suggest that existing universal visual tracking methods require further improvement to effectively tackle the challenges posed by increasingly complex transformation scenarios.
These unsatisfactory tracking results underscore the need for further enhancement in existing visual tracking methods to effectively address the challenges pose by transforming objects.

% \begin{figure}[htbp]
%   \centering
%   \subfigure{\includegraphics[width=0.49\textwidth]{LaTeX2e_Proceedings_Templates/figs/prec.png}}
%   \hfill
%   \subfigure{\includegraphics[width=0.49\textwidth]{LaTeX2e_Proceedings_Templates/figs/succ.png}}
%   \caption{11}
%   \label{fig:figure_label}
% \end{figure}

% \begin{figure}[htbp]
%   \centering
%   \subfigure{\includegraphics[width=0.3\textwidth]{LaTeX2e_Proceedings_Templates/figs/SV.PNG}}
%   \subfigure{\includegraphics[width=0.3\textwidth]{LaTeX2e_Proceedings_Templates/figs/SV.png}}

%   \subfigure{\includegraphics[width=0.3\textwidth]{LaTeX2e_Proceedings_Templates/figs/SV.png}}
%   \caption{11}
%   \label{fig:figure_label}
% \end{figure}

\section{Conclusion}

In this paper, we delve into a novel visual object tracking challenge: tracking transforming objects.  
Specifically, we present the Dataset for Tracking Transforming Objects (DTTO), which is the first benchmark dedicated to this task.
Furthermore, to evaluate the performance of existing trackers and establish a baseline for future comparisons, we conduct a comprehensive evaluation of 20 state-of-the-art tracking algorithms, accompanied by in-depth analysis. 
Our experimental results emphasize the importance of refining and advancing current visual tracking methodologies to effectively tackle the challenges posed by transforming objects. 
We believe that the proposed DTTO will facilitate future research and applications concerning tracking transforming objects. 
% This advancement is particularly valuable for more advanced applications like video surveillance and motion analysis, where understanding the changes in an object's behavior or state is crucial.


\begin{thebibliography}{8}

\bibitem{cui2022mixformer}
Y.~Cui, et~al.: Mixformer: End-to-end tracking with iterative mixed attention. In: 2022 CVPR, pp. 13608--13618 (2022).

\bibitem{Li2020AsymmetricDC}
S.~Li, et~al.: Asymmetric discriminative correlation filters for visual tracking. Frontiers Inf. Technol. Electron. Eng., vol.~21, pp. 1467--1484 (2020).

\bibitem{Li2021LearningRC}
S.~Li, et~al.: Learning residue-aware correlation filters and refining scale estimates with the grabcut for real-time uav tracking. In: 2021 3DV, pp. 1238--1248 (2021).

\bibitem{LI2022108614}
S.~Li, et~al.: Learning residue-aware correlation filters and refining scale for real-time uav tracking, PR, p. 108614 (2022).

\bibitem{ye2022joint}
B.~Ye, et~al.: Joint feature learning and relation modeling for tracking: A one-stream framework. In: 2022 ECCV, pp. 341--357 (2022).

\bibitem{gao2023generalized}
S.~Gao, et~al.: Generalized relation modeling for transformer tracking. In: 2023 CVPR, pp. 18686--18695 (2023).

% \bibitem{zhong2023fisher}
% P.~Zhong, et~al.: Fisher pruning for developing real-time uav trackers. Journal of Real-Time Image Processing, vol.~20, no.~5, p.~91 (2023).

\bibitem{li2023adaptive}
S.~Li, et~al.: Adaptive and background-aware vision transformer for real-time uav tracking. In: 2023 ICCV, pp. 13989--14000 (2023).

\bibitem{zeng2023towards}
D.~Zeng, et~al.: Towards discriminative representations with contrastive instances for real-time uav tracking. In: 2023 ICME, pp. 1349--1354 (2023).

\bibitem{wang2023learning}
X.~Wang, et~al.: Learning disentangled representation with mutual information maximization for real-time uav tracking. In: 2023 ICME, pp. 1331--1336 (2023).


\bibitem{zhang2020ocean}
Z.~Zhang, et~al.: Ocean: Object-aware anchor-free tracking. In: 2020 ECCV, pp. 771--787 (2020).

\bibitem{guo2020siamcar}
D.~Guo, et~al.: Siamcar: Siamese fully convolutional classification and regression for visual tracking. In: 2020 CVPR, pp. 6269--6277 (2020).

\bibitem{danelljan2020probabilistic}
M.~Danelljan, et~al.: Probabilistic regression for visual tracking. In: 2020 CVPR, pp. 7183--7192 (2020).

\bibitem{yan2021learning}
B.~Yan, et~al.: Learning spatio-temporal transformer for visual tracking. In: ICCV, 2021, pp. 10448--10457.

\bibitem{Jiao2021DeepLI}
L.~Jiao, et~al.: Deep learning in visual tracking: A review. IEEE TNNLS, vol.~PP (2021).

\bibitem{wu2023dropmae}
Q.~Wu, et~al.: Dropmae: Masked autoencoders with spatial-attention dropout for tracking tasks. In: 2023 CVPR , pp. 14561--14571 (2023).

% \bibitem{shi2024explicit}
% L.~Shi, et~al.: Explicit visual prompts for visual object tracking. In: 2024 AAAI, 2024.

\bibitem{shi2024explicit}
L.~Shi, et~al.: Explicit visual prompts for visual object tracking. In: 2024 AAAI, vol.~38, no.~5, pp. 4838--4846 (2024).


% \bibitem{xie2024autoregressive}
% J.~Xie, et~al.: Autoregressive queries for adaptive tracking with spatio-temporal  transformers, CVPR, 2024.

\bibitem{xie2024autoregressive}
J.~Xie, et~al.: Autoregressive queries for adaptive tracking with spatio-temporal transformers. In: 2024 CVPR, pp. 19300--19309 (CVPR).


\bibitem{lilearning}
Y.~Li, et~al.: Learning adaptive and view-invariant vision transformer for real-time uav tracking. In: 2024 ICML (2024).

\bibitem{wang2024enhancing}
X.~Wang, , et~al.: Enhancing uav tracking: a focus on discriminative representations using contrastive instances. Journal of Real-Time Image Processing, vol.~21, no.~3, p.~78 (2024).


\bibitem{tokmakov2023breaking}
P.~Tokmakov, et~al.: Breaking the" object" in video object segmentation. In: 2023 CVPR, pp. 22836--22845 (2023).

\bibitem{yu2023video}
J.~Yu, et~al.: Video state-changing object segmentation. In: 2023 ICCV, pp. 20439--20448 (2023).

\bibitem{goyal2023m3t}
R.~Goyal, et~al.: M3t: Multi-scale memory matching for video object segmentation and tracking. arXiv preprint arXiv:2312.08514 (2023).

\bibitem{malim1994cognitive}
T.~Malim, Cognitive processes: attention, perception, memory, thinking and language. Bloomsbury Publishing (1994) .

\bibitem{estes2022handbook}
W.~Estes, Handbook of learning and cognitive processes. Psychology Press (2022).

% \bibitem{von1995cognitive}
% B.~Von~Eckardt, What is cognitive science? MIT press, 1995.

\bibitem{busemeyer2010cognitive}
J.~R. Busemeyer and A.~Diederich, Cognitive modeling. Sage (2010).


\bibitem{thatcher2021foundations}
R.~W. Thatcher et~al.: Foundations of cognitive processes. Routledge (2021).

\bibitem{kellogg2003cognitive}
R.~T. Kellogg, Cognitive psychology. Sage, vol.~2 (2003).


% \bibitem{schank2013scripts}
% R.~C. Schank and R.~P. Abelson, \emph{Scripts, plans, goals, and understanding: An inquiry into human knowledge structures}.\hskip 1em plus 0.5em minus 0.4em\relax Psychology press, 2013.

% \bibitem{mangawati2018object}
% A.~Mangawati, M.~Leesan, H.~R. Aradhya \emph{et~al.}, ``Object tracking algorithms for video surveillance applications,'' in \emph{2018 ICCSP}.\hskip 1em plus 0.5em minus 0.4em\relax IEEE, 2018, pp. 0667--0671.

% \bibitem{sahithi2023enhancing}
% A.~Sahithi, \emph{et~al.}, ``Enhancing object detection and tracking from surveillance video camera using yolov8,'' in \emph{2023 ICRAIS}.\hskip 1em plus 0.5em minus 0.4em\relax IEEE, 2023, pp. 228--233.

% \bibitem{xiao2023motiontrack}
% C.~Xiao, \emph{et~al.}, ``Motiontrack: Learning motion predictor for multiple object tracking,'' \emph{arXiv preprint arXiv:2306.02585}, 2023.

% \bibitem{li2023motion}
% Y.~Li, L.~Wu, Y.~Chen, X.~Wang, G.~Yin, and Z.~Wang, ``Motion estimation and multi-stage association for tracking-by-detection,'' \emph{CAIS}, pp. 1--14, 2023.


\bibitem{huang2019got}
L.~Huang, et~al.: Got-10k: A large high-diversity benchmark for generic object tracking in the wild. TPAMI, vol.~43, no.~5, pp. 1562--1577 (2019).

\bibitem{fan2019lasot}
H.~Fan, et~al.: Lasot: A high-quality benchmark for large-scale single object tracking. In: 2019 CVPR, pp. 5374--5383 (2019).

\bibitem{muller2018trackingnet}
M.~Muller, et~al.: Trackingnet: A large-scale dataset and benchmark for object tracking in the wild. In: 2018 ECCV, pp. 300--317 (2018).

\bibitem{wu2013online}
Y.~Wu, et~al.: Online object tracking: A benchmark. In: 2013 CVPR, pp. 2411--2418 (2013).

\bibitem{wu2015otb}
Y. ~Wu, et~al.: Object tracking benchmark. TPAMI, vol.~37, no.~9, pp. 1834--1848 (2015).

\bibitem{kristan2016novel}
M.~Kristan, et~al.: A novel performance evaluation methodology for single-target trackers. TPAMI, vol.~38, no.~11, pp. 2137--2155 (2016).

\bibitem{kiani2017need}
H.~Kiani~Galoogahi, et~al.: Need for speed: A benchmark for higher frame rate object tracking. In: 2017 ICCV, pp. 1125--1134 (2017).


\bibitem{fan2021lasot}
H.~Fan, et~al.: Lasot: A high-quality large-scale single object tracking benchmark. IJCV, vol. 129, pp. 439--461 (2021).

\bibitem{liang2015encoding}
P.~Liang, et~al.: Encoding color information for visual tracking: Algorithms and benchmark. T-IP, vol.~24, no.~12, pp. 5630--5644 (2015).

\bibitem{li2015nus}
A.~Li, et~al.: Nus-pro: A new visual tracking challenge. TPAMI, vol.~38, no.~2, pp. 335--349 (2015).

\bibitem{mueller2016benchmark}
M.~Mueller, et~al.: A benchmark and simulator for uav tracking. In: 2016 ECCV, pp. 445--461 (2016).

\bibitem{bertinetto2016fully}
L.~Bertinetto, et~al.: Fully-convolutional siamese networks for object tracking. In: 2016 ECCV, pp. 850--865 (2016).

\bibitem{fan2019siamese}
H.~Fan and H.~Ling.: Siamese cascaded region proposal networks for real-time visual tracking. In: 2019 CVPR, pp. 7952--7961 (2019).

\bibitem{fu2021stmtrack}
Z.~Fu, et~al.: Stmtrack: Template-free visual tracking with space-time memory networks. In: 2021 CVPR, pp. 13\,774--13\,783 (2021).


\bibitem{zhang2019deeper}
Z.~Zhang and H.~Peng.: Deeper and wider siamese networks for real-time visual tracking. In: 2019 CVPR, pp. 4591--4600 (2019).

\bibitem{dosovitskiy2020image}
A.~Dosovitskiy, et~al.: An image is worth 16x16 words: Transformers for image recognition at scale. In: 2020 ICLR (2020).

\bibitem{vaswani2017attention}
A.~Vaswani, et~al.: Attention is all you need. In: 2017 NeurIPS, vol.~30 (2017).

\bibitem{chen2021transformer}
X.~Chen, et~al.: Transformer tracking. In: 2021 CVPR, pp. 8126--8135 (2021).


\bibitem{cai2023robust}
Y.~Cai, et~al.: Robust object modeling for visual tracking. In: 2023 ICCV, pp. 9589--9600 (2023).

\bibitem{kou2024zoomtrack}
Y.~Kou, et~al.: Zoomtrack: Target-aware non-uniform resizing for efficient visual tracking. In: 2023 NeurIPS, vol.~36 (2023).

% \bibitem{alayrac2017joint}
% J.-B. Alayrac, I.~Laptev, J.~Sivic, and S.~Lacoste-Julien, ``Joint discovery of object states and manipulation actions,'' in \emph{ICCV}, 2017, pp. 2127--2136.

% \bibitem{souvcek2022look}
% T.~Sou{\v{c}}ek, \emph{et~al.}, ``Look for the change: Learning object states and state-modifying actions from untrimmed web videos,'' in \emph{CVPR}, 2022, pp. 13\,956--13\,966.

% \bibitem{grauman2022ego4d}
% K.~Grauman, \emph{et~al.}, ``Ego4d: Around the world in 3,000 hours of egocentric video,'' in \emph{CVPR}, 2022, pp. 18\,995--19\,012.

\bibitem{zhang2021learn}
Z.~Zhang, et~al.: Learn to match: Automatic matching network design for visual tracking. In: 2021 ICCV, pp. 13\,339--13\,348 (2021).

\bibitem{paul2022robust}
M.~Paul, et~al.: Robust visual tracking by segmentation. In 2022 ECCV, pp. 571--588 (2022).

\bibitem{mayer2022transforming}
C.~Mayer, et~al.: Transforming model prediction for tracking. In: 2022 CVPR, pp. 8731--8740 (2022).

\bibitem{chen2023seqtrack}
X.~Chen, et~al.: Seqtrack: Sequence to sequence learning for visual object tracking. In: 2023 CVPR, pp. 14\,572--14\,581 (2023).

\bibitem{kang2023exploring}
B.~Kang, et~al.: Exploring lightweight hierarchical vision transformers for efficient visual tracking. In: 2023 ICCV, pp. 9612--9621 (2023).

\bibitem{li2024lightweight}
Y.~Li, et~al.: Lightweight full-convolutional siamese tracker. KBS, vol. 286, p. 111439 (2024).

\bibitem{zhu2024dcpt}
J.~Zhu, et~al.: Dcpt: Darkness clue-prompted tracking in nighttime uavs. In: 2024 ICRA (2024).

\end{thebibliography}
%

\end{document}